\documentclass[10pt,twocolumn,letterpaper]{article}

\usepackage[pagenumbers]{cvpr} 

\usepackage[dvipsnames]{xcolor}

\definecolor{cvprblue}{rgb}{0.21,0.49,0.74}
\usepackage[all]{nowidow}
\usepackage[pagebackref,breaklinks,colorlinks,citecolor=cvprblue]{hyperref}

\def\paperID{21} 
\def\confName{CVPR}
\def\confYear{2024}

\title{NutritionVerse-Direct: Exploring Deep Neural Networks for Multitask Nutrition Prediction from Food Images}

\author{Matthew Keller\\
University of Waterloo\\
Waterloo, ON\\
{\tt\small m6keller@uwaterloo.ca}
\and Chi-en Amy Tai \\
University of Waterloo\\
Waterloo, ON\\
{\tt\small amy.tai@uwaterloo.ca}
\and Yuhao Chen \\
University of Waterloo\\
Waterloo, ON\\
{\tt\small yuhao.chen1@uwaterloo.ca}
\and Pengcheng Xi \\
National Research Council Canada\\
Ottawa, ON\\
{\tt\small pengcheng.xi@nrc-cnrc.gc.ca}
\and Alexander Wong \\
University of Waterloo\\
Waterloo, ON\\
{\tt\small alexander.wong@uwaterloo.ca}
}

\begin{document}
\maketitle
\begin{abstract}
Many aging individuals encounter challenges in effectively tracking their dietary intake, exacerbating their susceptibility to nutrition-related health complications. Self-reporting methods are often inaccurate and suffer from substantial bias; however, leveraging intelligent prediction methods can automate and enhance precision in this process. Recent work has explored using computer vision prediction systems to predict nutritional information from food images. Still, these methods are often tailored to specific situations, require other inputs in addition to a food image, or do not provide comprehensive nutritional information.

This paper aims to enhance the efficacy of dietary intake estimation by leveraging various neural network architectures to directly predict a meal's nutritional content from its image. Through comprehensive experimentation and evaluation, we present NutritionVerse-Direct, a model utilizing a vision transformer base architecture with three fully connected layers that lead to five regression heads predicting calories (kcal), mass (g), protein (g), fat (g), and carbohydrates (g) present in a meal. NutritionVerse-Direct yields a combined mean average error score on the NutritionVerse-Real dataset of 412.6, an improvement of 25.5\% over the Inception-ResNet model, demonstrating its potential for improving dietary intake estimation accuracy.
\end{abstract}    
\section{Introduction}
\label{sec:intro}

Elderly populations often face significant challenges when it comes to dietary intake tracking, often exacerbated by health complications~\cite{ahmed2010assessment}. Additionally, the task of documenting daily food intake can be burdensome, requiring individuals to search for individual food items in databases and track overall nutrient consumption. Previous studies~\cite{pooled-results-from-val,structure-dietary-measurement} have shown that employing self-tracking methods suffers from significant bias. Consequently, there is a growing need to automate the process of tracking nutrition intake, alleviating the burden of manual alternatives. Leveraging smartphone cameras for nutrition tracking presents a promising solution, with deep neural networks offering a potential avenue for automation. 

A recent dataset that was released was NutritionVerse-Real (NV-Real)~\cite{tai2023nutritionversereal}, an image dataset of meals from various camera angles. Using this dataset, another study~\cite{tai2023nutritionverseempirical} identified that direct predictions from images outperform an indirect approach that estimates nutritional information from a segmented image mask. In our investigation, we extend this groundwork by examining and modifying deep-learning architectures using the real-image dataset to iteratively redesign a baseline network for multitask nutrition prediction. Specifically, we modify a multitask deep-learning architecture to predict calorie, mass, protein, carbohydrate, and fat information from a food image. Our objective is to investigate the efficacy of employing deep neural networks to analyze food images for nutritional content estimation.

In particular, our study seeks to address two key questions: 
\begin{enumerate}
    \item Does modifying the model architecture's fully connected layers contribute to improved nutrition prediction?
    \item Does the utilization of a vision transformer or masked autoencoder model enhance dietary intake estimation performance? 
\end{enumerate}

Our motivation behind exploring different fully connected layer structures is to leverage the pre-trained weights of our base feature extractor to make dietary intake estimations. We aim to exploit latent features from these layers without having to learn many additional weights of fully connected layers to explore their efficacy in capturing latent features relating to dietary intake information.

Similarly, we seek to leverage other image-processing architectures, notably vision image transformers (ViT)~\cite{vit-wu} and masked autoencoders (M-AutoE)~\cite{he2021masked} to explore their ability to capture image latent features related to food intake estimation.

Through systematic experimentation and evaluation of the NV-Real dataset, we aim to provide insights into the effectiveness of these architectural modifications for enhancing the accuracy of nutrition estimation from food images.

\section{Related Work}
\label{sec:related_work}

Nutrition prediction from food images has garnered attention in previous studies, but often involves additional inputs alongside RGB food images or relies on datasets with limited diversity of camera viewpoints.

DepthCalorieCam \cite{depth-calorie-cam} showcases volume-based calorie prediction methods; however, its performance is demonstrated only for a few food types and does not aim to predict other nutritional information. Computer vision-based calorie estimation proposed in \cite{cv-based-calorie} leverages depth information to predict nutritional information, but also only explores calculating caloric content and is limited to individual food items rather than full meals. Methods such as those proposed by SimFoodLoc \cite{sim-food-loc} and NutriNet \cite{nutrinet} focus on classifying food items present in a meal, but do not predict nutritional information from these images.

Geolocation-based approaches, such as Menu-Match \cite{menu-match} and Im2Calories \cite{im-2-calories}, utilize location and image inputs to match food to menu items in nearby restaurants. However, their applicability is limited to restaurant food and relies on information of a user's location. Additionally, the accuracy of nutritional content determination is not detailed.

\begin{figure}[t]
  \centering
  \includegraphics[width=\linewidth]{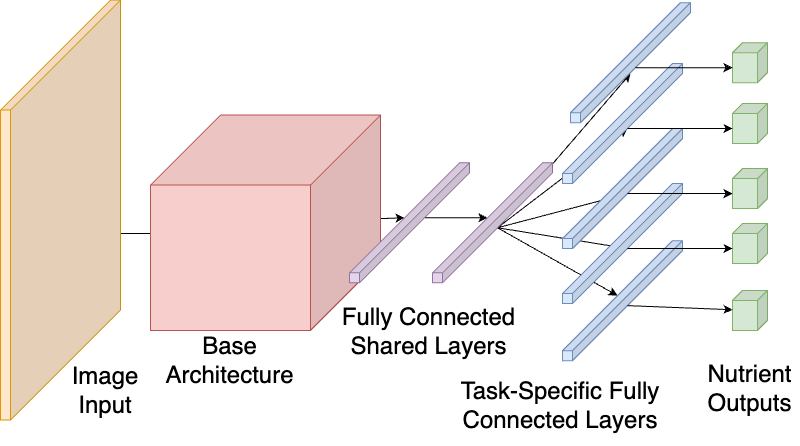}
  \caption{Architecture used for nutrient prediction.}
  \label{fig:full-architecture}
\end{figure}

Nutrition5k \cite{nutrition5k}, on the other hand, introduces a dataset and proposes a deep learning architecture for detecting nutritional information from image inputs using a multitask regression model. While the dataset includes a wide variety of food, it represents only a few unique camera angles. Moreover, only one method for direct prediction is proposed, highlighting the need for further exploration of deep learning architectures for predicting various nutritional components from images.

\section{Methods}
\label{sec:methods}

\begin{figure}[t]
  \centering
  \includegraphics[width=0.9\linewidth]{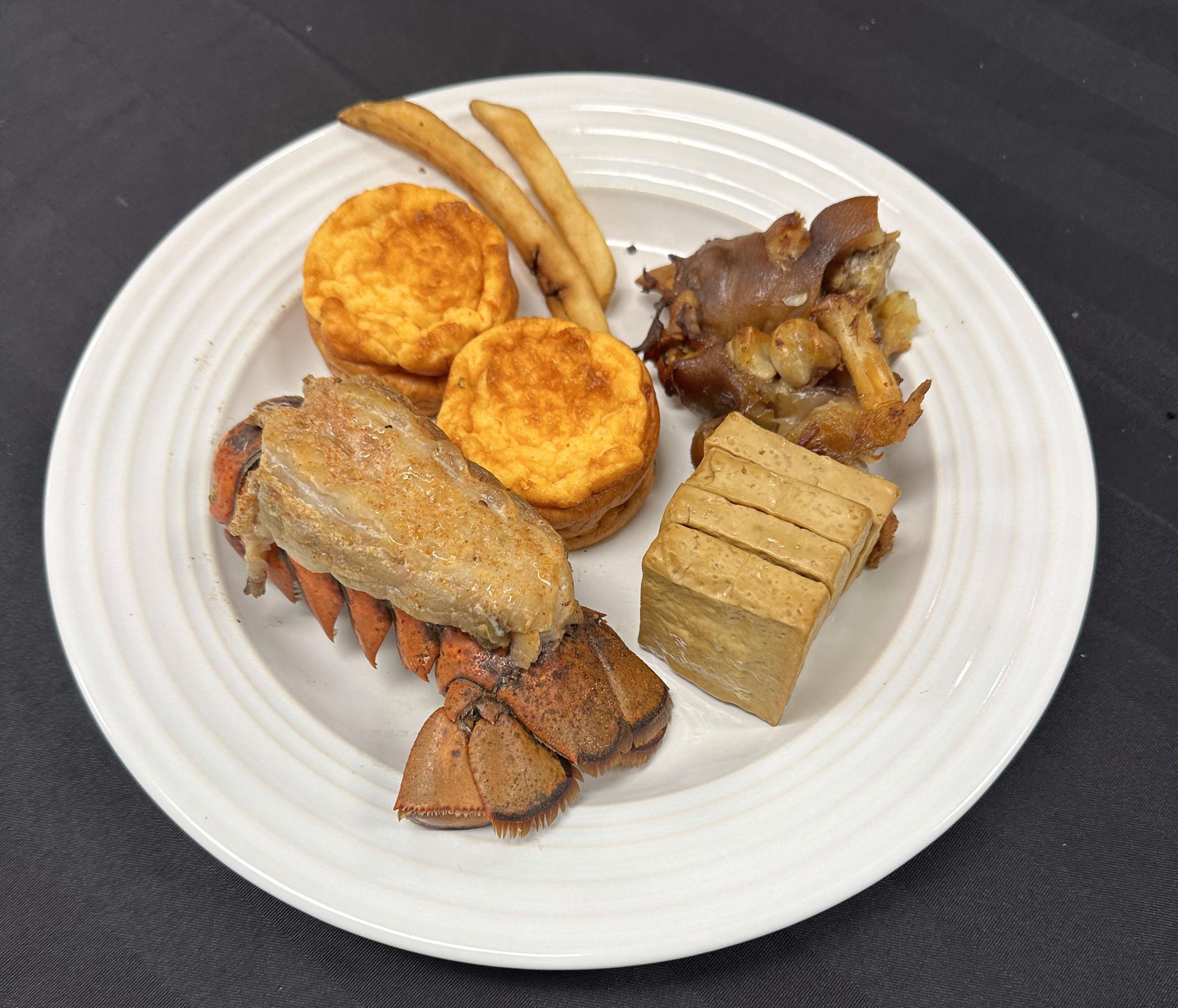}
  \caption{Food meal example image of a from NV-Real~\cite{tai2023nutritionversereal}.}
  \label{fig:nv-real-example}
\end{figure}

We use a deep learning architecture to predict calorie, mass, carbohydrate, fat, and protein content comprised of a feature extractor base and fully connected layers shared between tasks and task-specific fully connected layers. We adopt the architectural framework outlined in~\cite{tai2023nutritionverseempirical}, which is built upon the foundation established by~\cite{nutrition5k}. 

\begin{table*}[h]
  \centering
  \begin{tabular}{@{}lccccccc@{}}
    \toprule
    Base & Compressed & Calorie (kcal) & Mass (g) & Protein (g) & Fat (g) & Carb (g) & Combined \\
    \midrule
    Inception-Resnet & No & 356.3 & 123.7 & 26.9 & 18.5 & 28.5 & 554.0 \\
    Inception-Resnet & Yes & 305.5 & 162.1 & 35.2 & 17.4 & 51.3 & 571.5 \\
    ViT & No & \textbf{253.7} & \textbf{98.4} & \textbf{22.1} & \textbf{14.3} & \textbf{24.2} & \textbf{412.6} \\
    ViT & Yes & 311.9 & 149.5 & 27.3 & 16.2 & 46.8 & 551.7 \\
    M-AutoE & No & 463.3 & 144.3 & 32.0 & 24.4 & 53.5 & 717.5 \\
    M-AutoE & Yes & 476.2 & 152.0 & 30.6 & 22.9 & 56.6 & 737.3 \\
    \bottomrule
  \end{tabular}
  \caption{Model performance using compressed architecture vs full architecture with the best result bolded.}
  \label{tab:results-breakdown}
\end{table*}

Within this structure seen in Figure~\ref{fig:full-architecture}, the outputs of the feature extractor cascade into fully connected layers. We begin with using an Inception-ResNet~\cite{inception-resnet} base feature extractor pre-trained on ImageNet~\cite{vit-wu} and later explore using a ViT ~\cite{vit-wu} and M-AutoE ~\cite{he2021masked} as the feature extractor layer.

The outputs of the last hidden layer of the base architecture are passed into a series of two fully connected layers whose weights are shared across all five tasks. After, a task-specific fully connected layer followed by a linear output are used to gather the predicted content for each regression task - calories, mass, protein, fat, and carbohydrates. 

We leverage the NV-Real dataset with an example image shown in Figure~\ref{fig:nv-real-example} to train various models with RGB images with no normalization applied to color channels. During each experiment, we used an RMSProp optimizer with a learning rate set to 0.0001, epsilon set to 1.0, weight decay set to 0.9, and momentum set to 0.9. Both ViT and M-AutoE models were comprised of 12 attention heads and 12 hidden layers, and the M-AutoE decoder consisted of 16 attention heads and 8 hidden layers. All models were trained with a batch size of 32, and we employed a mean absolute error (MAE) loss function described in Equation~\ref{eq:mae}, where \( N \) represents the number of data points, \( y_i \) denotes the true value of the \( i \)-th data point, and \( \hat{y}_i \) represents the predicted value for the \( i \)-th label. The MAE loss is calculated by taking the absolute difference between each predicted value and its corresponding true value, summing these absolute differences over all labels, and then averaging them.

\begin{equation}\label{eq:mae}
\text{MAE} = \frac{1}{N}\sum_{i=1}^{N} |y_{i} - \hat{y}_{i}|
\end{equation}

\section{Experiments}
\label{sec:experiments}

The investigation into different architectures aims to optimize model performance by minimizing the combined MAE of all nutrient tasks. By comparing the efficacy of various feature extractors and fully connected layer configurations, we seek to identify architectures that strike a balance between accuracy and resource utilization, facilitating the widespread adoption of our solution for dietary tracking applications.

To determine the architecture, different changes were made to the original base architecture and their performance was compared across models for each regression task. First, we investigated changing the fully connected layers appended to our base architecture, then explored changing the base feature extractor with a vision image transformer and masked autoencoder architecture.

\subsection{What is the Best Fully Connected Layer Architecture?}

\begin{figure}[t]
  \centering
  \includegraphics[width=\linewidth]{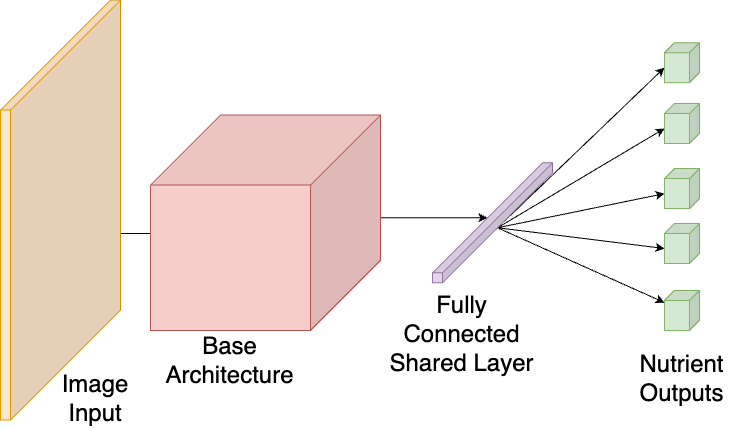}
  \caption{Compressed architecture with fewer fully connected layers.}
  \label{fig:compressed-architecture}
\end{figure}

Firstly, we introduce a modification to the fully connected layer section of our architecture, employing a single shared fully connected layer that feeds into separate regression heads, as depicted in Figure~\ref{fig:compressed-architecture}. This adjustment, referred to as the compressed architecture, aims to maximize the utilization of the pre-trained weights from the ImageNet~\cite{imagenet-deng} dataset within the base feature extractor architecture, thereby predicting each nutrient task without the need to train additional fully connected layers. As illustrated in Table~\ref{tab:results-breakdown}, the implementation of the non-compressed architecture resulted in improved performance, indicated by a lower combined MAE across the five tasks.

While the full architecture, constructed upon the Inception-ResNet base feature extractor, exhibited a lower combined MAE across all tasks compared to the compressed fully connected layers, we note that the compressed model showcased a notable reduction in MAE specifically for the calorie prediction task. This observation, coupled with the marginal disparity in the combined MAE between both predictors, prompted further exploration of the compressed architecture in subsequent experiments.

\subsection{What is the Best Feature Extractor Layer in Direct Prediction?}

Next, we address improving model performance by replacing the base feature extractor layers with a ViT architecture~\cite{vit-wu} a M-AutoE~\cite{he2021masked} to explore different architectures' abilities to capture food image features that contribute to nutritional information. Each of these architectures was also pre-trained on ImageNet~\cite{imagenet-deng} and used to replace the feature extractor in the complete and compressed fully connected layer set-ups shown in Figure~\ref{fig:full-architecture} and Figure~\ref{fig:compressed-architecture}, respectively. 

As shown in Table~\ref{tab:results-breakdown}, the best-performing feature extractor overall was the ViT, where either model trained with this as the base layer achieved a combined MAE score lower than all other models tested, suggesting this architecture is the most capable of capturing hidden features of food images contributing to nutritional information. Notably, the combined MAE for ViT was 412.6, an improvement of 25.5\% compared to the Inception-ResNet model.

The low performance of the M-AutoE model can be attributed to the relatively large hidden feature outputs of the model, which limited the downstream layers' capacity to effectively leverage the extracted features. This may have resulted in some of the rich feature representations to have not been preserved in the predictor layers, diminishing the model's ability to accurately capture and utilize crucial information for making predictions.

\section{Conclusion}
\label{sec:conclusion}

This paper highlights that the ViT architecture with two fully connected layers between tasks, a task-specific fully connected layer, and a prediction head with no removed fully connected layers performed the best compared to Inception-ResNet and M-AutoE. Compared to the next best base model (Inception-ResNet), ViT achieved a combined MAE that was better than the next best by over 25.5\%. 

Future work to further optimize the performance and efficiency of the dietary prediction model includes addressing magnitude differences between tasks by using different loss functions or normalizing task labels. Additionally, leveraging other pre-trained weights for the feature extractor base such as Nutrition5k~\cite{nutrition5k} or NutritionVerse-Synthetic~\cite{tai2023nutritionverse3d} can be explored.

{
    \small
    \bibliographystyle{ieeenat_fullname}
    \bibliography{main}
}


\end{document}